\documentclass[runningheads]{llncs}
\usepackage{graphicx}
\usepackage{multirow}
\usepackage{amsmath}
\usepackage[dvipsnames]{xcolor}
\usepackage{hyperref}
\usepackage{array}
\graphicspath{figs}
\newcommand{\cooltitle}{Efficient Bayesian Uncertainty Estimation \\for nnU-Net}
\newcommand{\coolshorttitle}{Efficient Bayesian Uncertainty Estimation for nnU-Net}

\newcommand{\simpauth}[1]{#1 \textit{et al.}}

\newcommand{\mymod}[2]{#1\ \text{mod}\ #2}
\newcommand{\bx}{\mathbf{x}}
\newcommand{\by}{\mathbf{y}}
\newcommand{\bw}{\mathbf{w}}

\newcommand{\bW}{\mathbf{W}}
\newcommand{\cd}{\mathcal{D}}
\newcommand{\percent}[1]{$#1\%$}

\newcommand{\myeqref}[1]{Eqn. (#1)}
\newcolumntype{P}[1]{>{\centering\arraybackslash}p{#1}}
\newcolumntype{M}[1]{>{\centering\arraybackslash}m{#1}}

\begin{document}
\title{\cooltitle}
\titlerunning{\coolshorttitle}
\author{Yidong Zhao\inst{1} \and
    Changchun Yang\inst{1} \and 
    Artur Schweidtmann\inst{2} \and
    Qian Tao\inst{1}}
%index{Zhao, Yidong}
%index{Yang, Changchun}
%index{Schweidtmann, Artur}
%index{Tao, Qian}

\authorrunning{Y. Zhao, C. Yang, A. Schweidtmann, Q. Tao}
\institute{Department of Imaging Physics, Delft University of Technology
\and Department of Chemical Engineering, Delft University of Technology \\
    \email{q.tao@tudelft.nl}}
\maketitle              % typeset the header of the contribution
\begin{abstract}
The self-configuring nnU-Net has achieved leading performance in a large range of medical image segmentation challenges. It is widely considered as the model of choice and a strong baseline for medical image segmentation. However, despite its extraordinary performance, nnU-Net does not supply a measure of uncertainty to indicate its possible failure. This can be problematic for large-scale image segmentation applications, where data are heterogeneous and nnU-Net may fail without notice. In this work, we introduce a novel method to estimate nnU-Net uncertainty for medical image segmentation. We propose a highly effective scheme for posterior sampling of weight space for Bayesian uncertainty estimation. Different from previous baseline methods such as Monte Carlo Dropout and mean-field Bayesian Neural Networks, our proposed method does not require a variational architecture and keeps the original nnU-Net architecture intact, thereby preserving its excellent performance and ease of use. Additionally, we boost the segmentation performance over the original nnU-Net via marginalizing multi-modal posterior models. We applied our method on the public ACDC and M\&M datasets of cardiac MRI and demonstrated improved uncertainty estimation over a range of baseline methods. The proposed method further strengthens nnU-Net for medical image segmentation in terms of both segmentation accuracy and quality control\footnote{Code is available at \url{https://gitlab.tudelft.nl/yidongzhao/hmc_uncertainty}.}.

\end{abstract}
\section{Introduction}
% =======================  Non-sense background
Manually delineating the region of interest from medical images is immensely expensive in clinical applications. Among various automatic medical image segmentation methods, the U-Net architecture~\cite{ronneberger2015u}, in particular the self-configuring nnU-Net framework~\cite{isensee2021nnu}, has achieved state-of-the-art performance in a wide range of medical image segmentation tasks~\cite{bernard2018deep,campello2021multi}. Nevertheless, the reliability of neural networks remains a major concern for medical applications, due to their over-parametrization and poor interpretability. Using the softmax output as a categorical probability distribution proxy is known to cause model miscalibration, which often leads to over-confidence in errors~\cite{mehrtash2020confidence}. This can be problematic for large-scale medical image segmentation applications, where data are heterogeneous and nnU-Net may fail without notice. Accurate uncertainty estimation is highly important in clinical deployment of automatic segmentation models~\cite{jungo2019assessing}.

% =======================  Related works: Uncertainty estimation 
Previous efforts on uncertainty estimation of neural networks are categorized as either Bayesian or non-Bayesian~\cite{mehrtash2020confidence}. Several non-Bayesian strategies have been proposed to quantify the predictive uncertainty in medical image segmentation. \simpauth{Guo} proposed to use temperature scaling~\cite{guo2017calibration} to mitigate the miscalibration in softmax. Probabilistic U-Net~\cite{kohl2018probabilistic} aims at generating a set of feasible segmentation maps. \simpauth{Baumgartner} proposed a hierarchical model to learn the distribution in latent spaces and make probabilistic predictions~\cite{baumgartner2019phiseg}. However, these methods require modification of the original network, sometimes to a significant extent, therefore hard to be integrated with the well-configured nnU-Net.

Unlike non-Bayesian approaches, Bayesian neural networks (BNNs) learn the weight posterior distribution given the training data. At inference time, a probabilistic prediction is made, which delivers a natural epistemic uncertainty~\cite{blundell2015weight}. However, deriving weight posterior is an intractable problem, and Variational Inference (VI)~\cite{blei2017variational} is often used for its approximation. The popular mean-field VI~\cite{blundell2015weight} doubles the network parameter assuming independence among weights, and can be unstable during training~\cite{ovadia2019can}. Gal \textit{et al.} proposed Monte-Carlo Dropout (MC-Dropout)~\cite{gal2016dropout} as a VI proxy by enabling dropout layers at test time. The method is theoretically and practically sound, however, recent works~\cite{folgoc2021mc,gonzalez2021self,gonzalez2021detecting} reported silent failure and poor calibration, with degraded segmentation performance. Deep Ensembles~\cite{lakshminarayanan2016simple} estimate network uncertainty via averaging independently trained networks. It is robust~\cite{ovadia2019can} but highly costly in computation, especially for models which demand lengthy training themselves.

% ====================== a teaser figure
\begin{figure}[t]
    \centering
    \includegraphics[width=0.96\textwidth]{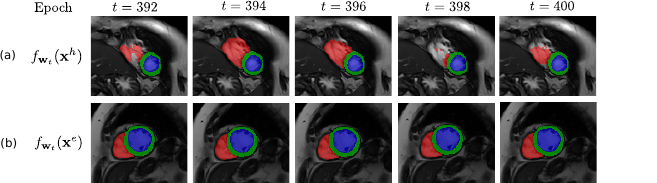}
    \caption{We observe that the network checkpoints at various training epochs make diverse predictions when the network is uncertain on a \textit{hard} input $\bx^h$ (a). On the contrary, predictions of good quality on an \textit{easy} input $\bx^e$ enjoys consistency across checkpoints (b). We leverage this phenomenon to perform Bayesian inference and quantify uncertainty of network predictions.}%\qian{meaning of upper-script $u$ and $c$ are not clear, what does this refer to}
    \label{fig:intro_figure}
\end{figure}

% =======================  Our method
We propose a novel method for medical image segmentation uncertainty estimation, which substantially outperforms the popular MC-Dropout, while being significantly more efficient than deep ensembles. We are inspired by the optimization theory that during stochastic gradient descent (SGD), the network weights continuously explore the solution space, which is approximately equivalent to weight space posterior sampling~\cite{mandt2017stochastic,maddox2019simple,mingard2021sgd}. While taken at appropriate moments of SGD, the network can be sampled \emph{a posteriori}, and uncertainty can be reflected in the agreement among these posterior models, as illustrated in Fig. \ref{fig:intro_figure}. This generic methodology enables VI approximation on any network, including the delicately-configured nnU-Net. The following contributions are made:
\begin{itemize}
    \item[-] A novel VI approximation method that realizes efficient posterior estimation of a deep model. 
    \item[-] An extension to the nnU-Net for uncertainty estimation in medical image segmentation, with significantly improved performance over several baseline methods including MC-Dropout.
    \item[-] Further improvement of the segmentation performance beyond the original nnU-Net, as showcased on cardiac magnetic resonance (CMR) data.
\end{itemize}

%To promote the diversity of weight samples, we further use the cyclical training strategy~\cite{huang2017snapshot} to encourage the training trajectory visit multiple posterior modes in the weight space. Combining checkpoints in different training cycles, we efficiently perform a multi-modal Bayesian inference without any modification on the network structure.\\

%via averaging checkpoints in the network training process using stochastic gradient descent (SGD) with or without momentum. 

\section{Methods}

\subsection{Bayesian Inference}
\label{sec:vi}
Given a training dataset $\cd=\left\{(\bx_i, \by_i)\right\}_{i=1}^N$ with $N$ image and segmentation label pairs, the BNN fits the posterior $p(\bw \vert \cd)$ over network weights $\bw$. At inference time, the probabilistic prediction on a test image $\bx^*$ can be formulated as the marginalization over parameter $\bw$:
\begin{equation}
    p(\by^*\vert \bx^*, \cd) = \int p(\by^*| \bx^*, \bw) p(\bw \vert \cd)\ d\bw
    \label{eq:bayesian_inf_int}
\end{equation}
We describe the proposed trajectory-based posterior sampling from $p(\bw \vert \cd)$ in Sec.~\ref{sec:sgd_vi} and our final multi-modal Bayesian inference based on cyclical training in Sec.~\ref{sec:multimodal}.

\subsection{SGD Bayesian Inference}
\label{sec:sgd_vi}
\subsubsection{Single-modal Posterior Sampling} During the SGD training, the weights first move towards a local optimum and then oscillate in the vicinity. The standard nnU-Net uses the polynomial learning rate decay scheme, in which the learning rate approaches zero as the epoch increases. The vanishing learning rate causes the network weights to converge. In order to make the weights explore a wider neighborhood around the local optimum, we follow~\cite{izmailov2018averaging} to set the learning rate to a constant value after $\gamma$ fraction of the total epoch budget $T$. For each training epoch $t \in [\gamma T, T]$, the weight checkpoint $\bw_t$ is approximately a sample drawn from a single-modal weight posterior: $\bw_t \sim p(\bw|\cd)$. In Fig.~\ref{fig:intro_figure} (a) and (b), we show the exemplar predictions made by various checkpoints on an \textit{easy} and a \textit{hard} example, respectively.

\subsubsection{Inference and Uncertainty Estimation} The $p(\by^*| \bx^*, \bw)$ term in \myeqref{\ref{eq:bayesian_inf_int}} corresponds to the model forward pass $f_{\bw}(\bx)$ at weights $\bw$. However, both the weight posterior and the integral are intractable for deep neural networks. In this work, we keep the intermediate weights (i.e. checkpoints) during SGD training as posterior samples after the learning stabilizes: $\bW=\{\bw_t | \gamma T<t\leq T\}$. Then, the Monte-Carlo approximation for the predictive posterior $p(\by^*\vert \bx^*, \cd)$ can be computed by taking the discrete version of~\myeqref{\ref{eq:bayesian_inf_int}}: 
\begin{equation}
    p(\by^*\vert \bx^*, \cd) \approx \frac{1}{n} \sum_{i=1}^{n} p(\by^*| \bx^*, \bw_{t_i})
    \label{eq:bayesian_inf_mc}
\end{equation}
where $\bw_{t_i}\in \bW$ and $n$ is the number of checkpoints out of the posterior sampling of the model. To generate the uncertainty map, we compute the entropy of the predictive $C-$class categorical distribution $H(\by^*_{i, j})$ for each voxel $(i, j)$, with 0 indicating the lowest uncertainty.
\begin{equation}
    H(\by^*_{i, j}) = -\sum_{k=1}^C p(\by^*_{i,j}=k\vert \bx^*, \cd) \log_2 p(\by^*_{i,j}=k\vert \bx^*, \cd)
    \label{eq:bayesian_uncertainty_ent}
\end{equation}

\subsubsection{Stochastic Weight Averaging} An alternative to the prediction space averaging in \myeqref{\ref{eq:bayesian_inf_mc}} is the Stochastic Weight Averaging (SWA)~\cite{izmailov2018averaging}, which takes average in the weight space instead: $p(\by^*\vert \bx^*, \cd) = p(\by^*| \bx^*, \overline{\bw})$, where $\overline{\bw}=\frac{1}{n} \sum_{i=1}^{n} \bw_{t_i}$, to construct a new aggregated model with improved performance~\cite{izmailov2018averaging}. After weight averaging, the network needs to perform one more epoch of forward passes to update the batch normalization layer parameters before inference.

\subsection{Multi-modal Posterior Sampling}
\label{sec:multimodal}
\subsubsection{Cyclical Learning Rate} The traditional SGD converges to the neighborhood of a single local optimum, hence the checkpoints capture only a single mode of the weight posterior. The diversity in weight samples has proven beneficial for uncertainty estimation~\cite{fort2019deep,fuchs2021practical}. In order to capture a multi-modal geometry in weight space, we employ a cyclical learning rate~\cite{huang2017snapshot} which periodically drives the weights out of the attraction region of a local optimum. However, the cosine annealing scheme proposed in~\cite{huang2017snapshot} causes instability in the nnU-Net training. We divide the training epoch budget $T$ to $M$ cycles, each of which consumes $T_c=\frac{T}{M}$ epochs. A high restart learning rate $\alpha_r$ is set only for the first epoch in each training cycle. After $\gamma$ fraction of $T_c$ epochs, we keep the gradient update step constant. Our proposed learning rate scheme is formulated as \myeqref{\ref{eq:cyclical_lr}}:
\begin{equation}
    \alpha(t) =
    \begin{cases}
        \alpha_r &, t_c  = 0\\
        \alpha_0\left[1-\frac{\min(t_c,\gamma T_c)}{T}\right]^{\epsilon} &, t_c  > 0
    \end{cases}
    \label{eq:cyclical_lr}
\end{equation}
where $t_c = \mymod{t}{T_c}$ is the in-cycle epoch number and the constant exponent $\epsilon$ controls the decaying rate of the learning rate .
\subsubsection{Multi-modal Checkpoint Ensemble} In the $c^{th}$ training cycle, we collect checkpoints $\bW_c=\{\bw_t|\gamma T_c \leq \mymod{t}{T_c} \leq T_c-1\}$.Then the aggregated checkpoints of all $M$ training cycles $\bW = \cup_{c=1}^M \bW_c$ consist of multi-modal posterior samples in the weight space. We build an ensemble of $n$ models via sampling $\frac{n}{M}$ intermediate weights from each mode $\bW_c$ for $c\in\{1, 2, \dots, M\}$. The multi-modal Bayesian inference is then performed as in~\myeqref{\ref{eq:bayesian_inf_mc}}. In comparison to~\cite{huang2017snapshot}, which collected one snapshot in each cycle, our method combines both local and global uncertainty of the weights.

\section{Experiments}
\subsection{Experimental Setup}
\subsubsection{Datasets} We trained our network on the publicly available ACDC dataset\footnote{\url{https://www.creatis.insa-lyon.fr/Challenge/acdc}}~\cite{bernard2018deep}, which contains annotated end-diastolic (ED) and end-systolic (ES) volumes of 100 subjects. We randomly split $80\%$ of The ACDC data for network training and kept the rest $20\%$ to validate the in-domain (ID) performance. To evaluate the performance under a domain shift, i.e. out-of-domain (OOD), we also tested our method on the M\&M dataset\footnote{\url{https://www.ub.edu/mnms}}~\cite{campello2021multi} collected from different MRI vendors and medical centers. The annotated part includes 75 subjects collected from Vendor A (Siemens) in a single medical center and 75 scans by Vendor B (Phillips) from two medical centers.

\subsubsection{Implementation Details} For CMR segmentation, we trained the 2D nnU-Net, with the proposed learning rate modulation in Sec.~\ref{sec:sgd_vi} and \ref{sec:multimodal}. During each training cycle, SGD with monentum was used for optimization with a batch size of 20. We set the initial learning rate as $\alpha_0=0.01$, the decay exponent $\epsilon=0.9$ and the restart learning as $\alpha_r=0.1$. We trained $M=3$ cycles within a total training budget of $T=1,200$ epochs, which is slightly higher than the standard nnU-Net which consumes $1,000$ epochs for a single model training. The learning rate kept at a constant level after $\gamma=80\%$ of the epochs in each individual training cycle. We chose the latest $n=30$ checkpoints (epoch Step $2$) in the last training cycle to build a single-modal checkpoints ensemble (\emph{Single-modal Ckpt. Ens.}). For a multi-modal weight sampling (\emph{Multi-modal Ckpt. Ens.}), we aggregated the latest $10$ checkpoints of all three training cycles. We took the single model prediction at the end of the training as the baseline (\emph{Vanilla}). Temperature scaling (\emph{Temp. Scaling}) with $\tau=1.5$ was implemented as a calibration baseline as in~\cite{guo2017calibration}. We also compared our proposed method to MC-Dropout~\cite{gal2016dropout} and deep ensemble~\cite{lakshminarayanan2016simple} (\emph{Deep Ens.}). The dropout probability was set as $p=0.1$ and $n=30$ predictions were drawn for uncertainty estimation. A 5-model deep ensemble was trained on the ACDC training data with random initialization. We also evaluated the performance of \emph{SWA}~\cite{izmailov2018averaging}, which averages the posterior weights sampled during the last training cycle, and we used the softmax predictions for uncertainty estimation.

\subsubsection{Evaluation Metrics}
On each dataset, the segmentation performance was evaluated using the mean Dice coefficients of the three foreground classes, namely, left ventricle (LV), right ventricle (RV), and myocardium (MYO), over all volumes. We quantify the calibration performance through the expected calibration error (ECE) of voxels in the bounding box of all foreground classes in each volume~\cite{guo2017calibration,mehrtash2020confidence}. We divided the voxels into 100 bins according to their confidence from \percent{0} to \percent{100}. The ECE is defined as the weighted average of the confidence-accuracy difference in each bin: $
    \mathrm{ECE} = \sum_{s=1}^{100} \frac{|B_s|}{N_v}\left\vert \mathrm{conf}(B_s) - \mathrm{acc}(B_s)\right\vert$,
where $B_s$ defines the set of voxels whose confidence falls in the range $\left(\left(s-1\right)\%, s\%\right]$, $\mathrm{conf}(B_s)$ is the mean confidence of voxels in bin $B_s$, $\mathrm{acc}(B_s)$ represents the classification accuracy of the voxels in $B_s$ and $N_v$ is the total number of voxels.
\subsection{Results}

% ===================== t-SNE plot
\begin{figure}[t]
    \centering
    \includegraphics{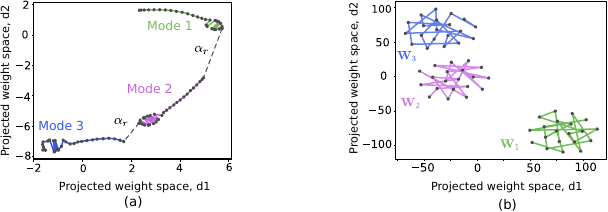}
    \caption{(a) t-SNE plot of the weight space during SGD training. Dotted lines illustrate the transition between weight modes. (b) t-SNE plot of the posterior weights, which bounce in the neighborhood of different modes.}
    \label{fig:fig_tsne}
\end{figure}

\subsubsection{Training Trajectory Visualization} We first present the training trajectory in the weight space to validate our concept of multi-modal posterior. Fig.~\ref{fig:fig_tsne}(a) shows the t-SNE plot of the weights of the last decoder layer in nnU-Net during the three training cycles. We observe that the cyclical training triggered a long-range movement in the weight space such that multiple local optima were visited. In Fig.~\ref{fig:fig_tsne}(b) we demonstrate both local and global weight posterior distribution. In each training cycle, the weights did not converge to a single point but kept bouncing in a local neighborhood, conforming to~\cite{mandt2017stochastic,maddox2019simple}. The diversity in posterior modes is clearly shown by the three clusters in the t-SNE weight space. We note that similar behavior of multi-modality was also observed for posterior weights in other layers. 
\begin{table}[h]
\centering
\caption{Dice coefficients on ID and OOD test sets.}
\label{tab:dice}
\begin{tabular}{|M{0.13\textwidth}|M{0.09\textwidth}M{0.09\textwidth}M{0.09\textwidth}|M{0.08\textwidth}M{0.09\textwidth}M{0.09\textwidth}|M{0.09\textwidth}M{0.09\textwidth}M{0.09\textwidth}|}
\hline
\multirow{2}{*}{Method} & \multicolumn{3}{c|}{ACDC Validation (ID)} & \multicolumn{3}{c|}{M\&M Vendor A (OOD)} & \multicolumn{3}{c|}{M\&M Vendor B (OOD)} \\ \cline{2-10}
    & RV & MYO & LV & RV & MYO & LV & RV & MYO & LV\\ \hline\hline
    Vanilla & 0.911 $\pm0.052$& 0.904 $\pm0.026$ & 0.944 $\pm0.035$ & 0.830 $\pm0.119$ & 0.810 $\pm0.043$ & 0.897 $\pm0.063$ & 0.878 $\pm0.077$ & 0.844 $\pm0.054$ & 0.894 $\pm0.076$ \\ \hline
    SWA & 0.913 $\pm0.054$  & 0.910 $\pm0.023$  & 0.948 $\pm0.033$  & 0.856 $\pm0.093$ & 0.808 $\pm0.041$ & 0.896 $\pm0.061$ & 0.879 $\pm0.067$ & 0.845 $\pm0.050$ & 0.897 $\pm0.069$  \\ \hline
    MC-Dropout& 0.906 $\pm0.061$ & 0.901 $\pm0.028$ & 0.940 $\pm0.043$ & 0.810 $\pm0.146$  & 0.810 $\pm0.049$ & 0.896 $\pm0.066$ & 0.880 $\pm0.079$ & 0.844 $\pm0.058$ & 0.891 $\pm0.081$ \\ \hline
    Deep Ens. & 0.915 $\pm0.051$ & 0.912 $\pm0.023$ & \textbf{0.951} $\pm0.029$  & \textbf{0.857} $\pm0.089$  & 0.816 $\pm0.041$  & 0.902 $\pm0.062$   & \textbf{0.885} $\pm0.068$  & 0.849 $\pm0.047$ & 0.897 $\pm0.064$ \\ \hline
    Ckpt. Ens. (Single) & \textbf{0.919} $\pm0.051$  & 0.912 $\pm0.022$  & \textbf{0.951} $\pm0.032$ & 0.851 $\pm0.100$ & 0.816 $\pm0.041$ & 0.902 $\pm0.061$ & 0.883 $\pm0.070$ & 0.850 $\pm0.048$ & \textbf{0.901} $\pm0.065$ \\ \hline
    Ckpt. Ens. (Multi) & 0.918 $\pm0.051$ & \textbf{0.913} $\pm0.024$ & \textbf{0.951} $\pm0.031$ & 0.852 $\pm0.105$ & \textbf{0.818} $\pm0.043$ & \textbf{0.905} $\pm0.060$ & \textbf{0.885} $\pm0.069$ & \textbf{0.851} $\pm0.050$ & 0.899 $\pm0.071$ \\
\hline
\end{tabular}
\end{table}

\subsubsection{Segmentation and Calibration Performance} We list the average Dice coefficient of all methods in comparison in Table~\ref{tab:dice}. The table demonstrates that except for MC-Dropout, which downgraded the performance slightly, the rest methods improved the segmentation performance in comparison to the Vanilla nnU-Net. The proposed checkpoint ensembles achieved overall the best performance except for the RV on Vendor A.

\begin{table}[h]
\centering
\caption{ECE ($\%$) on ID and OOD test sets.}\label{tab:ece_bbox}
\begin{tabular}{|c|P{0.23\textwidth}|P{0.23\textwidth}|P{0.23\textwidth}|}
\hline
    Methods & ACDC Validation (ID)& M\&M Vendor A (OOD)& M\&M Vendor B (OOD)\\ \hline\hline
    Vanilla & 2.56 & 4.34 & 3.79\\
    Temp. Scaling & 2.18 & 3.91 & 3.46\\
    SWA & 2.39 & 4.07 &  3.70\\
    MC-Dropout & 1.70 & 3.41 &  2.95\\
    Deep Ens. & 1.63  & 3.16 &  2.85\\
    % SWAG & 1.80 & 3.48 & 3.11\\
    % Multi-modal SWAG  & 1.57 & 3.19 & 2.86\\ 
    Ckpt. Ens. (Single) & \textbf{1.25} & 2.83 &  2.69\\
    Ckpt. Ens. (Multi) & \textbf{1.25} & \textbf{2.75} & \textbf{2.61} \\\hline
\end{tabular}
\end{table}

The ECE scores are listed in Table~\ref{tab:ece_bbox}. From the table, we can observe that the temperature scaling slightly improved the calibration performance. The results of Vanilla and SWA show that a single model was poorly calibrated compared to all 4 Bayesian methods. MC-Dropout improved calibration but was not as effective as Deep Ensemble. The proposed single-model checkpoint ensemble achieved the best performance on ID data, and the multi-modal checkpoint ensemble further improved the performance on OOD data.

\subsubsection{Qualitative Results}
\begin{figure}[t]
    \centering
    \includegraphics[width=0.96\textwidth]{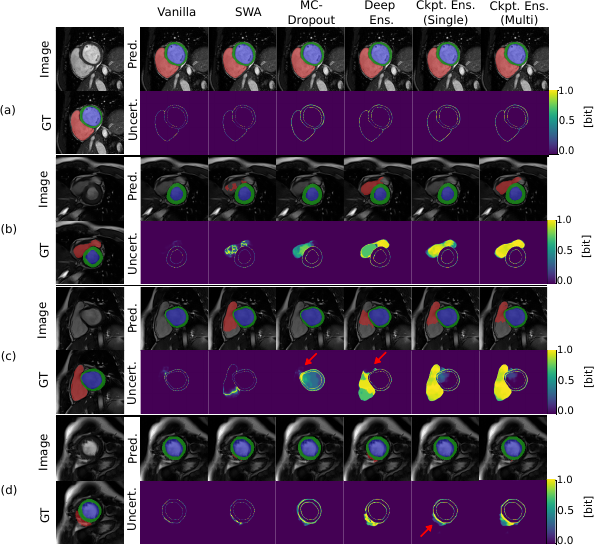}
    \caption{Predictions (Pred.) and estimated uncertainty maps (Uncert.) on a successful case (a) and three partially failed cases (b-d). In case (a) all the methods highlighted the border as uncertain. MC-Dropout failed in all the three cases (b-d) to delineate RV, while reporting low uncertainty in the corresponding area. Deep Ensemble is robust but missed part of the uncertain areas in case (c). Multi-modal weight sampling detected the failed RV area more robustly than the single-modal version in case (d).}
    \label{fig:qualitative}
\end{figure}

Some representative examples of the segmentation and uncertainty maps are shown in Fig.~\ref{fig:qualitative}. For the successful segmentation in Fig.~\ref{fig:qualitative}(a), all methods highlighted the prediction borders as uncertain compared to regions, which conforms to our intuition in practice that different observers may have slightly different border delineations. Fig.~\ref{fig:qualitative}(b-d) show three cases in which the network failed, mainly in the RV region. In practice, this is the most common failure even with the state-of-the-art nnU-Net. We can observe that MC-Dropout frequently output low uncertainty in the area of wrong predictions (RV). Deep Ensemble correctly identified the uncertain areas in case (b) and (d). However, the limited number of models in Deep Ensemble may miss part of the error, as shown in Fig.~\ref{fig:qualitative}(c) (red arrow). We further demonstrate the benefit of multi-modality in weight space through Fig.~\ref{fig:qualitative}(d) which shows an apical slice with a small RV. In this case, only deep ensemble and the multi-modal checkpoints ensemble successfully detected the uncertainty in RV region, while other methods just output high confidence, indicating ``silent" failure, i.e. mistake that will escape notice in automatic segmentation pipelines. 

In all three challenging cases, our proposed method robustly detected the nnU-Net failure on RV. Interestingly, we note that the estimated uncertainty map strongly correlated with true RV area, which is highly beneficial and informative for further manual contour correction in clinical practice.

\section{Conclusion}
In this work, we proposed an efficient Bayesian inference approximation method for nnU-Net, from the SGD training perspective. Our method does not require adapting the nnU-Net configuration and is highly efficient in computation. Our experimental results on both in-domain and out-of-domain cardiac MRI data proved its effectiveness, in comparison with established baseline methods such as MC-Dropout and Deep Ensemble. The proposed uncertainty estimation further strengthens nnU-Net for practical applications of medical image segmentation, in terms of both \emph{uncertainty estimation} and \emph{segmentation performance}. 

\section{Acknowledgement}
The authors gratefully acknowledge TU Delft AI Initiative for financial support. 

%The proposed method robustly estimated the uncertainty when the network failed to work. And the uncertainty maps are highly informative for the manual intervention to correct the network failure in clinical practice. 
%============================= qualitative results

\clearpage
\bibliographystyle{splncs04}
\bibliography{paper2172.bib}

\end{document}